\documentclass[times, review, 10pt]{elsarticle}

\usepackage{amssymb}
\usepackage{amsmath}
\usepackage{subfiles}
\usepackage{booktabs}
\usepackage{multirow}
\usepackage{rotating}
\usepackage{adjustbox}
\usepackage{pifont} 
\usepackage{soul}
\usepackage{algorithm}
\usepackage{algorithmic}
\usepackage[table,dvipsnames]{xcolor}
\usepackage{caption}
\usepackage{xcolor}
\usepackage{color}

\newcommand*\colourcheck[1]{%
  \expandafter\newcommand\csname #1check\endcsname{\textcolor{#1}{\ding{52}}}%
}

\colorlet{tdcolor}{yellow!35}
\sethlcolor{tdcolor}

\journal{Pattern Recognition}

\begin{document}

\begin{frontmatter}

%% Title, authors and addresses

\title{Thinking Before Retrieving: Robust Zero-Shot Composed Image Retrieval via Strategic Planning and Self-Criticism}
% PEC-Agent: A Planning–Execution–Criticism Framework for Training-Free Zero-Shot Composed Image Retrieval

\author[label1]{Gunho Jung\fnref{fn1}}
\ead{gh\_jung@korea.ac.kr}

\author[label1]{Jeong-Woo Park\fnref{fn1}}
\ead{jeongwoo_park@korea.ac.kr}

\author[label1]{Seon Bin Kim}
\ead{s_b_kim@korea.ac.kr}
         
\author[label1]{Seong-Whan Lee\corref{cor1}}
\ead{sw.lee@korea.ac.kr}

%% Author affiliation
\affiliation[label1]{organization={Department of Artificial Intelligence, Korea University},%Department and Organization
            city={Seoul},
            postcode={02841}, 
            country={Republic of Korea}}
            
% \affiliation[label2]{organization={Department of Brain and Cognitive Engineering, Korea University},%Department and Organization
%             city={Seoul},
%             postcode={02841}, 
%             country={Republic of Korea}}
\fntext[fn1]{Equal contribution.}
\cortext[cor1]{Corresponding author.}
%% Abstract
\begin{abstract}
Composed image retrieval requires identifying a target image from a gallery by integrating a reference image with a textual modification instruction. 
In a training-free zero-shot setting, this task relies on constructing a retrieval-oriented textual query within a frozen vision--language embedding space at inference time. 
Existing approaches predominantly rely on a single-pass generation strategy that fuses the reference context and modification text into a unified description. 
This strategy makes it difficult to detect or correct semantic distortions and omissions during generation.
Consequently, the preservation of reference attributes and the integration of textual requirements interfere with each other, which degrades retrieval precision.
To address these challenges, we introduce PEC-CIR, a training-free framework that structures query construction as a multi-stage reasoning pipeline. 
The framework operates through a Planner--Executor--Critic architecture where the Planner extracts explicit constraints, the Executor generates multiple candidate target descriptions, and the Critic evaluates these candidates based on constraint compliance. 
By reframing query construction as a staged inference process instead of a single-pass output, PEC-CIR reduces the propagation of generative errors by explicitly evaluating candidate queries before retrieval, thereby improving retrieval stability.

\end{abstract}

%% Keywords

\end{frontmatter}

%% main text
%%
%%%%%%%%%%%%%%%%%%%%% Introduction %%%%%%%%%%%%%%%%%%%%%
\section{Introduction}
Composed Image Retrieval (CIR) aims to retrieve a target image by synergizing a reference image with a textual instruction \cite{philbin2007object, gordo2016deep, lee2012pill, sun2026benchcir}.
Unlike conventional retrieval based solely on visual similarity, this task demands the preservation of specific visual attributes while simultaneously modifying features in accordance with a textual prompt \cite{wang2025generative, xing2025context}.
For example, in e-commerce, a query such as ``the same design but in a different color without a logo'' exemplifies the need for precise multimodal integration.
This cross-modal synergy enables the expression of fine-grained preferences, allowing for the retrieval of images that satisfy nuanced modifications while maintaining visual identity \cite{kang2014nighttime,delmas2022artemis,ju2025mire}.  
Compositional retrieval requirements arise across various application domains, including fashion, interior design, and personalized recommendation, highlighting the practical relevance of CIR in real-world scenarios.

\begin{figure*}[!t]
\centering
\includegraphics[width=\linewidth]{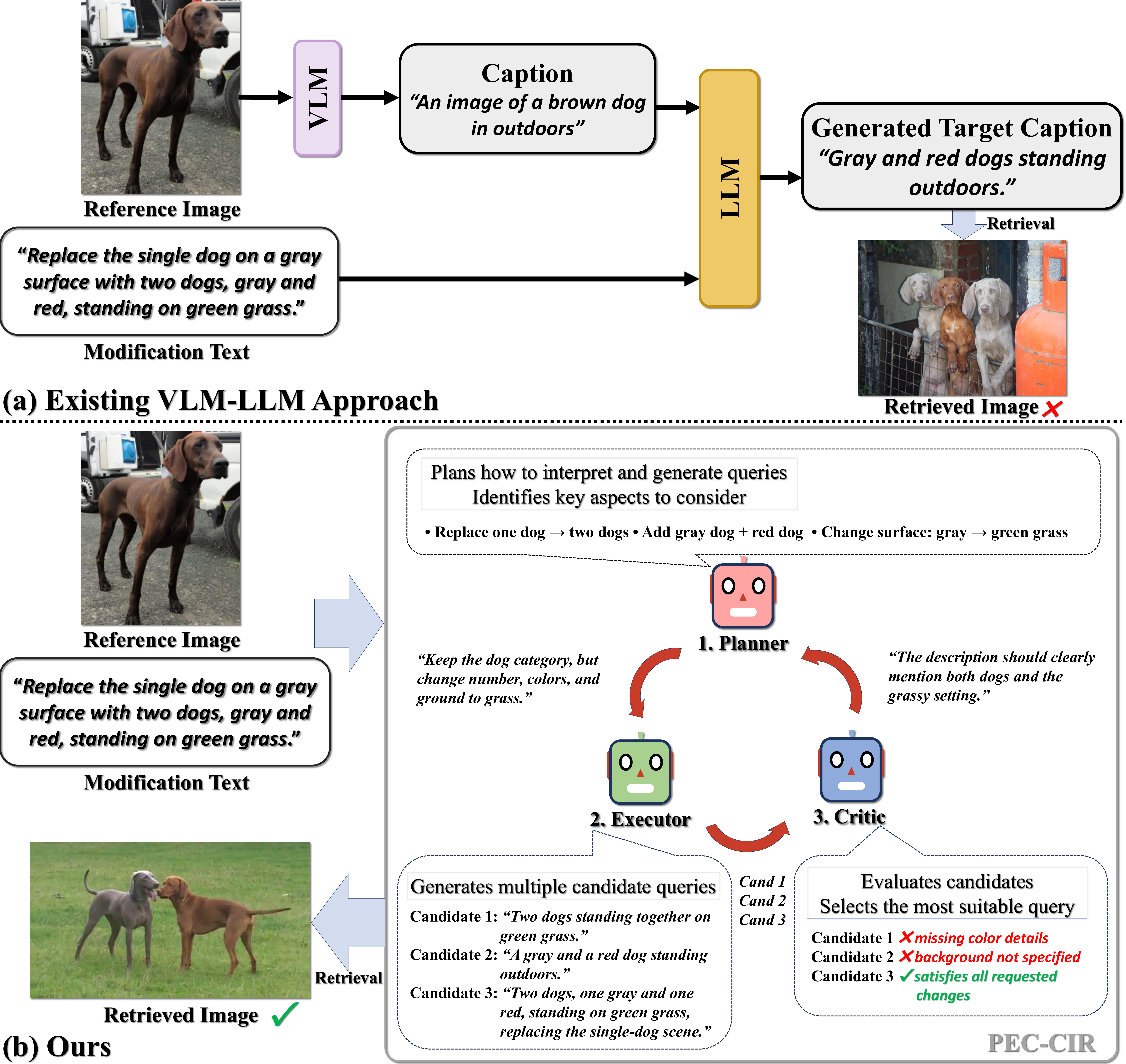}
\caption{Comparison between a typical training-free CIR approach and the proposed PEC-CIR.}
\label{fig:comparison}
\end{figure*}

Early CIR studies relied on supervised learning with annotated triplets consisting of a reference image, a modification text, and a target image \cite{baldrati2022effective, zhang2024magiclens}. 
While achieving strong performance through joint visual-textual representations, supervised frameworks require large-scale annotations and incur substantial computational costs. 
To minimize supervision overhead, recent work explores zero-shot CIR, leveraging pre-trained vision-language models to perform multimodal fusion without task-specific training \cite{agnolucci2025isearle,chen2026data,radford2021learning}. 
Notably, the integration of Large Language Models (LLMs) and Multimodal Large Language Models (MLLMs) emerges as a prominent paradigm owing to their advanced reasoning capabilities for translating complex multimodal instructions into descriptive target representations. 
The generative strategy produces a natural language description of the target image for subsequent text-to-image retrieval, facilitating a flexible and training-free inference process \cite{park2025mcot,tang2025reason,yang2024ldre, tu2025multimodal, cheng2025generative}.

However, existing generative zero-shot frameworks suffer from three structural limitations.
First, the decoupled processing of the reference image and textual instruction limits explicit cross-modal interaction between the visual context and the editing intent.
The separation creates ambiguity regarding the preservation or modification of specific visual attributes, resulting in inaccurate retrieval results.
Furthermore, the stochastic nature of text generation introduces semantic volatility during query construction.
Probabilistic decoding leads to the omission of critical visual constraints or the introduction of extraneous concepts, thereby undermining retrieval consistency. 
Finally, previous paradigms uncritically adopt generated outputs as optimal, overlooking the inherent volatility in retrieval quality across diverse candidates.
This absence of discriminative oversight allows hallucinations or semantic misalignments to propagate unchecked to the retrieval engine, compromising the robustness of training-free frameworks.
Figure \ref{fig:comparison}(a) illustrates a representative failure case of an existing training-free CIR approach, in which a single-pass generated query captures only partial aspects of the textual instruction, yielding misaligned retrieval results.

% 해결책
To address these challenges, we propose PEC-CIR, a robust training-free framework that redefines zero-shot query construction as a structured reasoning pipeline. 
Our framework decomposes the process into three interconnected stages: Planning, Execution, and Criticism.
The Planner first extracts structured constraints for preservation and modification, grounding the reasoning in visual context. The Executor then synthesizes diverse candidate descriptions to mitigate the volatility of stochastic generation. Finally, the Critic evaluates these candidates against objective criteria, ensuring that only descriptions satisfying strict retrieval requirements are propagated.
The multi-stage design stabilizes query construction, yielding consistent performance gains on the FashionIQ \cite{wu2021fashion} and CIRR \cite{liu2021image} benchmarks.

%기여

Our contributions are summarized as follows:
\begin{itemize}
    \item 
    We propose PEC-CIR, a training-free framework that redefines zero-shot CIR as a multi-stage reasoning pipeline, effectively mitigating the volatility inherent in single-pass generative models.
    \item 
    We introduce an agentic refinement mechanism facilitated by a feedback-driven loop, which prevents the propagation of hallucinations through rigorous visual-textual verification.
    \item 
    Experimental results on FashionIQ and CIRR benchmarks demonstrate that PEC-CIR achieves consistent performance gains, achieving state-of-the-art performance among training-free compositional retrieval methods.

\end{itemize}

%%%%%%%%%%%%%%%%%%%%% Related Works %%%%%%%%%%%%%%%%%%%%%
\section{Related Works}
\subsection{Composed Image Retrieval}
Composed Image Retrieval (CIR) aims to retrieve a target image from a gallery by integrating a reference image with a textual modification instruction \cite{philbin2007object, gordo2016deep, park2025far}. 
The fundamental difficulty entails identifying which visual attributes of the reference image require preservation and which necessitate modification according to the instruction. 
The dual requirements of preservation and modification represent the expression of complex editing intents and necessitate an intricate interaction between visual and textual modalities.
Early studies on CIR focused on supervised learning setting, using annotated triplets consisting of a reference image, a modification text, and labeled targets for training \cite{wen2023self, yang2023composed}. 
More broadly, supervised cross-modal retrieval has also explored modality-invariant representation learning and semantic relation modeling through label-aware graph convolution, dual adversarial graph neural networks, and multi-label contrastive learning  \cite{qian2022integrating, qian2021dual, qian2021adaptive}.
While effective within the distribution of annotated triplets, supervised designs inherently depend on large-scale supervision. Consequently, static architectures lack the flexibility to accommodate novel attributes or compositions without the significant overhead of additional data collection and model retraining \cite{chen2026data}. \nocite{lee1995multilayer, roh2010view, lee1990translation}
Related zero-shot cross-modal retrieval tasks have also been studied in zero-shot sketch-based image retrieval (ZS-SBIR), where models must align sketches and photos while generalizing to unseen categories. 
Recent ZS-SBIR studies address this challenge through relation-aware meta-learning, bidirectional knowledge distillation, and mixed-sample augmentation with bidirectional mining \cite{liu2025relation,du2025zero,liu2026zero}.

\subsection{Zero-Shot Composed Image Retrieval}
\subsubsection{Representation-based approach}
Representation-based frameworks focus on composing image and instruction features directly within the embedding space of pre-trained vision--language models \cite{baldrati2023zero, karthik2023vision, wu2024training, kim2025enhancing}. 
In vector-fusion paradigms, the reference image and the modification text undergo separate encoding before combining via weighted fusion \cite{karthik2023vision}, linear transformations \cite{byun2024reducing}, or similarity-based operations \cite{jang2024spherical}. 
Direct vector composition is computationally efficient and requires minimal task-specific training.
However, the reliance on pre-defined feature spaces limits the ability to represent complex semantics, particularly under conditional modifications where visual and textual contexts require deep integration \cite{chen2026data}.

\subsubsection{Mapping and Inversion-based approach}
Mapping-based strategies utilize pre-trained mapping modules to project visual features into the word embedding space, representing images as pseudo-word tokens for integration into textual prompts \cite{saito2023pic2word, wang2025mapping}.
Refined mapping frameworks further incorporate fine-grained visual details or external knowledge to enhance the descriptive expressivity of the generated tokens \cite{lin2024fine, suo2024knowledge}.
In contrast, Inversion-based approaches employ iterative optimization at inference time to identify textual embeddings that encapsulate the visual context of the reference image within the linguistic domain \cite{agnolucci2025isearle}.
However, reliance on pre-trained mapping modules or iterative optimization limits generalization to novel concepts and incurs substantial computational overhead.

\subsubsection{Generation-based approach}
Generation-based zero-shot CIR has recently emerged, utilizing LLMs or MLLMs to synthesize retrieval-oriented descriptions directly from multimodal inputs \cite{tang2025reason}.
Recent work further emphasizes the semantic editing process itself, where the relative text is treated as a semantic increment between the reference and target images for training-free zero-shot retrieval \cite{yang2024semantic}.
Generative pipelines rewrite instructions or generate modified captions, with the resulting text subsequently encoded as a retrieval query.
While offering enhanced expressive flexibility for complex modifications, single-pass generative methods remain highly sensitive to the volatility of a single output \cite{cheng2025generative}.
Directly propagating generative outputs to the retrieval stage enables stochastic errors to compromise retrieval performance.
Our method leverages a multi-stage reasoning agent that utilizes explicit planning and autonomous criticism to verify and refine the retrieval query.

\subsection{Agent-Based Reasoning Paradigm}
Agentic reasoning has emerged to enhance the problem-solving capabilities of LLM models by mitigating the constraints of single-step inference \cite{yao2022react,kojima2022large}. 
The architectures facilitate the decomposition of complex tasks into modular roles comprising planning, execution, and self-reflection to support iterative decision-making based on intermediate outcomes.
The iterative refinement of the outputs minimizes error accumulation, countering the risks associated with premature generation \cite{zubkova2026leveraging,zhou2022least}.

Within Composed Image Retrieval (CIR), the shift toward multi-stage inference reflects an effort to manage the stochasticity inherent in zero-shot generation \cite{tang2025reason}. 
Recent studies investigate staged procedures like instruction analysis or heuristic-based re-ranking to stabilize retrieval outcomes \cite{tu2025multimodal,wang2022self}.
Generative pipelines utilizing LLM for query synthesis permit initial generation errors to reach the retrieval stage without internal verification\cite{park2025mcot, cheng2025generative}.
To resolve the susceptibility to stochastic error propagation, PEC-CIR integrates explicit planning and autonomous criticism to verify and refine the retrieval query.

%%%%%%%%%%%%%%%%%%%%% Methodology %%%%%%%%%%%%%%%%%%%%%
\section{Methodology}
\label{sec:method}

\subsection{Problem Definition}
\label{subsec:problem}
Given a reference image $I_R$ and a modification text $T_M$, CIR aims to retrieve a target image $I_T$ from a gallery $\mathbf{G}=\{I_i\}_{i=1}^{N}$.
The retrieval objective entails the simultaneous preservation of the core visual identity of $I_R$ and the incorporation of semantic transformations specified in $T_M$.
In the training-free zero-shot setting, the framework leverages pre-trained vision-language models for image $\boldsymbol{f}_\theta$ and text encoder $\boldsymbol{f}_\phi$, respectively.
To bridge the gap between multimodal inputs and the retrieval space, the composed pair $(I_R, T_M)$ is projected onto a surrogate textual query $Q$.
The retrieval process is defined by the following function:
\begin{equation}
I^\ast = \arg\max_{I_i \in \mathbf{G}} \cos(\boldsymbol{f}_\theta(I_i), \boldsymbol{f}_\phi(Q)),
\end{equation}
where $\cos(\cdot, \cdot)$ denotes the cosine similarity between the latent representations.

\subsection{Overview of PEC-CIR}
\label{subsec:overview}
\begin{figure}[t]
\centering
\includegraphics[width=\linewidth]{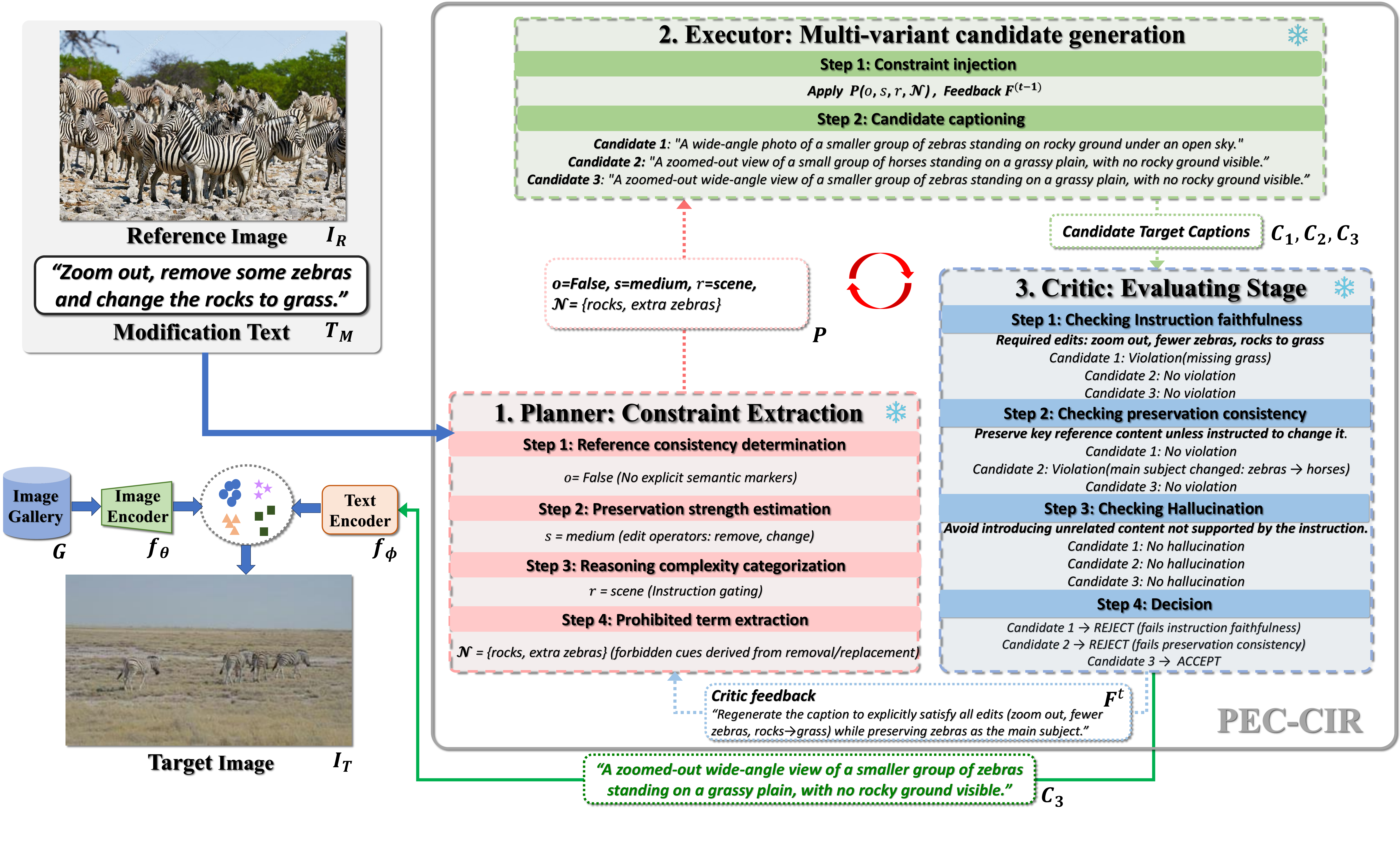}
\caption{Overview of PEC-CIR. The framework performs a Planner--Executor--Critic loop to iteratively generate and refine target descriptions, which are aggregated into the final query representation for composed image retrieval.}
\label{fig:pec_agent_framework}
\end{figure}

PEC-CIR employs a Planner--Executor--Critic framework to facilitate robust query construction in a zero-shot, training-free environment.
As illustrated in Figure~\ref{fig:pec_agent_framework}, our approach implements a systematic progression through three specialized modules designed to overcome the limitations of single-pass generation.
Initially, the Planner extracts structured constraints by jointly analyzing the reference image $I_R$ and the modification instruction $T_M$. 
These constraints define the boundaries for visual preservation and semantic transformation, ensuring the reasoning remains grounded in the visual context.
The Executor synthesizes a diverse set of candidate target descriptions on the extracted constraints.
This multi-candidate generation strategy expands the semantic coverage of the intended modification, allowing the framework to explore various linguistic expressions that capture the target visual attributes.
The Critic subsequently validates these candidates to ensure strict alignment with the reference context and transformation intent.
This stage enforces a rigorous verification mechanism, where the Critic either selects the most faithful candidate or triggers recursive refinement to eliminate potential hallucinations.
The finalized description is used as the surrogate textual query $Q$, which is projected into the latent space for retrieval.

\subsection{Planner: Constraint and Intent Analysis}
\label{subsec:planner}
Given the reference image $I_R$ and the modification instruction $T_M$, the Planner produces a plan $\boldsymbol{P}$ that guides candidate generation and evaluation. 
The plan encodes explicit control signals, including whether reference consistency is overridden, the degree of reference preservation, the complexity of the requested edit, and exclusion constraints derived from the instruction.
Formally, the Planner is defined as: 
\begin{equation}
\boldsymbol{P} = \mathcal{F}_{\mathrm{planner}}(I_R, T_M),
\end{equation}
where the resulting plan is represented as
\begin{equation}
\boldsymbol{P} = (o, s, r, \mathcal{N}).
\end{equation}
In this representation, $ o \in \{0, 1\}$ denotes whether reference consistency is overridden., $s \in \{\mathrm{high}, \mathrm{medium}, \mathrm{low}\}$ denotes the degree of reference preservation, $r \in \{\mathrm{object}, \mathrm{scene}, \mathrm{complex}\}$ denotes the reasoning granularity required by the instruction, and $\mathcal{N}$ denotes the set of prohibited terms derived from the instruction.
The Planner is implemented as an LLM-based reasoning agent, ensuring consistent and interpretable plan generation without the computational overhead of additional parameter optimization.

\subsubsection{Reference consistency determination}
The Planner first evaluates the modification instruction $T_M$ to identify explicit requests for visual deviation from the reference image $I_R$. 
Expressions, such as ``unlike the reference'' or ``different from the original'', indicate that the instruction prioritizes transformation over attribute preservation.
In such instances, the Planner assigns $o=1$, thereby overriding or relaxing reference consistency constraints during the subsequent evaluation phase. 
Conversely, in the absence of explicit deviation cues, the Planner assigns $o=0$, ensuring that the retrieval process maintains fidelity to the core visual identity of the reference image.

\subsubsection{Preservation strength estimation}
When reference consistency is not overridden $o=1$, the Planner estimates the preservation intensity $s$ by analyzing lexical cues within $T_M$. 
Invariance-oriented expressions such as ``same'', ``keep'', or ``unchanged'' signify a requirement for high visual fidelity, resulting in $s=\mathrm{high}$. 
Conversely, explicit modification operators including ``add'', ``remove'', or ``change'' signal a transformation-centric intent that necessitates visual grounding, yielding $s=\mathrm{medium}$. 
In the absence of triggers, the Planner assigns $\mathrm{low}$. 
This parameter modulates the weighting of reference consistency within the Critic's evaluation logic, ensuring a calibrated balance between preservation and modification.
When reference consistency is overridden $o=1$, the preservation term is relaxed or deactivated during Critic scoring.

\subsubsection{Reasoning complexity categorization}
Subsequently, the Planner assesses the reasoning granularity $r$ by utilizing the instruction length as a proxy for semantic complexity. 
Concise instructions, which typically target object-level or single-attribute modifications, are categorized as $r=\mathrm{object}$.
Instructions of intermediate length that encompass scene-wide context or multi-attribute combinations are assigned $r=\mathrm{scene}$, while extensive prompts involving multiple compositional requirements are designated as $r=\mathrm{complex}$.
This variable calibrates the strictness of the Critic’s faithfulness verification.
Specifically, it prevents the enforcement of overly rigid lexical coverage on complex instructions, where high-level semantic alignment is prioritized over literal token matching.
This adaptive gating mechanism ensures that the retrieval engine remains robust to the linguistic diversity of elaborate user requests.

\subsubsection{Prohibited term extraction}
Finally, the Planner constructs the prohibited term set $\mathcal{N}$ by isolating linguistic patterns of negation, removal, and substitution within $T_M$. 
This extraction process targets four primary categories: (i) lexical negations identified by tokens such as ``without'' or ``no'', (ii) explicit removal directives, (iii) contrastive exclusions signified by ``rather than'' or ``instead of'', and (iv) the antecedent $X$ within substitution patterns (e.g., ``replace $X$ with $Y$'' or ``change $X$ to $Y$''). The resulting set $\mathcal{N}$ enables the Critic to penalize violations of explicit exclusion constraints, effectively preventing the introduction of disallowed visual elements in the generated candidates.

In summary, the Planner distills the modification instruction into a suite of explicit control signals $\boldsymbol{P}=(o,s,r,\mathcal{N})$ that govern the entire reasoning pipeline.
Specifically, $o$ and $s$ calibrate the balance between preservation and modification across both generation and evaluation, while $r$ and $\mathcal{N}$ guide the content density and constraint satisfaction during the Executor's synthesis and the Critic's verification.

\subsection{Executor: Multi-Variant Candidate Generation}
\label{subsec:executor}
Guided by the structured plan $\boldsymbol{P}$, the Executor synthesizes a diverse set of candidate descriptions for the intended target image. The generated candidate set is defined as:
\[
\mathcal{C} = \{C_k\}_{k=1}^{K},
\]
where \(K\) denotes the number of generated variants and each $C_k$ represents a target description that incorporates the modification instruction while preserving relevant visual content from the reference image.
This module transcends simple lexical paraphrasing of $T_M$. 
Instead, it generates target representations specifically designed for semantic alignment with the frozen multimodal embedding space.
To mitigate the semantic volatility inherent in stochastic language models, the Executor produces multiple candidate variants, deferring the final selection to the Critic's discriminative evaluation.

\subsubsection{Constraint-aware synthesis}
The Executor integrates $(I_R, T_M, \boldsymbol{P})$ to generate candidate target descriptions that satisfy the constraints identified by the Planner. 
When $o=1$, the module deprioritizes reference preservation to emphasize instruction compliance, as the modification instruction explicitly requests deviation from the reference image.
Conversely, when $o=0$, the Executor preserves the core visual content of the reference image, with the preservation strength controlled by $s$. 
In particular, when $s=\mathrm{high}$, the Executor strengthens preservation constraints to ensure the consistent maintenance of the reference scene, primary entities, and unchanged visual attributes. 
Furthermore, the Executor incorporates the prohibited term set $\mathcal{N}$ as explicit negative constraints, preventing the introduction of disallowed elements in $C_k$.

\subsubsection{Multi-variant generation via diverse biases}
To maximize semantic coverage, the Executor employs multiple prompting strategies, each imposing a distinct inductive bias on the generated candidates.
Instead of relying on a single textual interpretation, we diversify the synthesis into three strategic directions consisting of (i) a conciseness-oriented approach that focuses strictly on the required modifications, (ii) a consistency-focused approach that prioritizes the continuity of the reference context, and (iii) an attribute-rich approach that encourages detailed descriptions of object properties and spatial relationships.
This multi-variant sampling ensures that the candidate set $C$ spans a broad spectrum of the visual-linguistic space, providing the Critic with a robust pool of potential queries for final selection.

\subsubsection{Feedback-driven execution}
Candidate refinement is driven by the feedback signal $F^{(t-1)}$ provided by the Critic, which the Executor leverages to rectify specific failure modes identified in the prior iteration.
This recursive process operates for a maximum of $T$ rounds, where each subsequent generation step $t$ is informed by the evaluative insights from step $t-1$ to systematically resolve semantic discrepancies or constraint violations.
The loop concludes once a candidate satisfies the Critic's verification criteria, ensuring that the final surrogate query is both faithful to the instruction and robust within the frozen embedding space.

\subsection{Critic: Multi-Step Verification and Selection}
\label{subsec:critic}
The Critic evaluates the candidate target descriptions generated by the Executor and identifies the candidate to be used in the final retrieval query. 
Its role is to verify whether each candidate satisfies the modification instruction, remains consistent with the reference content when required, and avoids prohibited or degenerate linguistic patterns. 
To support lexical comparison, let \(\tau(\cdot)\) denote an operator that maps a text sequence to the set of its salient lexical cues. 
The Critic then applies a sequence of verification steps followed by score-based selection.

\subsubsection{Lexical fidelity and instruction alignment}
The first step verifies whether a candidate preserves the modification cues specified in \(T_M\). For a candidate caption \(C_k\), lexical coverage is measured by the instruction recall:
\begin{equation}
\mathrm{Rec}(T_M, C_k)
=
\frac{|\tau(T_M)\cap \tau(C_k)|}{|\tau(T_M)|}.
\end{equation}
In parallel, semantic alignment with the modification instruction is measured by
\begin{equation}
s_{\mathrm{mod}}(k)=\cos\!\big(\boldsymbol{f}_\phi(T_M), \boldsymbol{f}_\phi(C_k)\big).
\end{equation}
The lexical recall score captures explicit cue preservation, whereas $s_{\mathrm{mod}}(k)$ accounts for semantic alignment beyond surface-level overlap. 
Both signals are incorporated into the Critic score, allowing the framework to balance literal instruction coverage with flexible semantic matching.

\subsubsection{Reference-aware preservation consistency}
The second step evaluates whether a candidate remains consistent with the reference image $I_R$ when preservation is required.
Reference consistency is measured as $s_{\mathrm{ref}}(k) = \cos(\boldsymbol{f}_\theta(I_R), \boldsymbol{f}_\phi(C_k))$, and its contribution is modulated by the planning signals $o$ and $s$ as:
\begin{equation}
w_{\mathrm{ref}}=
\begin{cases}
0, & \text{if} o = 1,\\
w_{s}, & \text{if} o = 0,
\end{cases}
\end{equation} 
where $w_{s} \in \{w_{high}, w_{med}, w_{low}\}$ is the coefficient associated with the preservation strength $s$. 
When $o = 1$, reference consistency is overridden, and the reference-consistency term is removed from the Critic score.
When $o = 0$, the Critic enforces reference consistency with a strength determined by $s$, thereby penalizing candidates that alter reference attributes not specified by the modification instruction.

\subsubsection{Linguistic and negative constraint compliance}
The third step filters candidates that exhibit excessive lexical dependence on the modification instruction or violate explicit negative constraints.
The overlap penalty $\Delta_{\mathrm{overlap}}$ is computed from the Jaccard similarity $\mathrm{Jac}(T_M, C_k)$ to discourage direct copying of the modification text.
The prohibition penalty $\Delta_{\mathrm{prohibit}}$ is applied when $C_k$ contains terms from the prohibited set $\mathcal{N}$ or when the Critic detects hallucinated content that is not supported by the visual-textual evidence.

\subsubsection{Multi-criteria scoring and acceptance}
Following the verification steps, the Critic integrates the evaluative signals into a composite score:
\begin{equation}
\begin{aligned}
S_k &= \alpha \cdot \mathrm{Rec}{(T_M, C_k)} + \beta \cdot s_{\mathrm{mod}}(k) + w_{\mathrm{ref}} \cdot s_{\mathrm{ref}}(k) \\
& + \Delta_{\mathrm{len}}(C_k) - \Delta_{\mathrm{overlap}}(T_M, C_k) - \Delta_{\mathrm{prohibit}}(\mathcal{N}, C_k),
\end{aligned}
\label{eq:critic_score}
\end{equation}
where $\alpha$ and $\beta$ are weighting coefficients, and $\Delta_{\mathrm{len}}(C_k)$ adjusts for target caption length.
The optimal candidate is identified by $k^{\ast} = \arg\max_{k}S_{k}$.
The selected candidate is accepted as the final surrogate query when it satisfies the Critic's reliability criteria.
Candidates with low composite scores or explicit constraint violations are regarded as unreliable, prompting the Critic to initiate the refinement loop with structured feedback.

\subsubsection{Feedback generation and refinement loop}
The Critic generates a structured feedback signal \(F^{(t)}\) for the next iteration once the selected candidate is regarded as unreliable.
The feedback specifies missing modification cues, reference-preservation issues, prohibited-term violations, and excessive lexical overlap with the original instruction.
This feedback guides the Executor to revise the next set of candidate target descriptions by emphasizing omitted constraints and suppressing detected errors.

The refinement follows the update rule:
\begin{equation}
\mathcal{C}^{(t+1)} \leftarrow \mathcal{F}_{\mathrm{Executor}}(I_R, T_M, \boldsymbol{P}, F^{(t)}),
\end{equation}
where $F^{(t)}$ provides corrective guidance derived from the Critic's evaluation.
This closed-loop process terminates once a reliable candidate is accepted or the maximum number of iterations $T$ is reached.

\subsection{Inference Procedure of PEC-CIR}
\label{subsec:inference_procedure}

\begin{algorithm}[!htbp]
\caption{Inference Procedure of PEC-CIR}
\label{alg:pec_cir}
\begin{algorithmic}[1]

\STATE \textbf{Input:} Reference image $I_R$, modification text $T_M$, gallery $\mathbf{G}$, image encoder $\boldsymbol{f}_\theta$, text encoder $\boldsymbol{f}_\phi$
\STATE \textbf{Parameters:} Maximum number of iterations $T$, number of candidates $K$
\STATE \textbf{Output:} Ranked gallery images

\STATE Obtain plan $\boldsymbol{P} = \mathcal{F}_{\mathrm{planner}}(I_R, T_M)$
\STATE Initialize feedback \(F^{(0)} \leftarrow \emptyset\)
\STATE Initialize final query $Q \leftarrow \emptyset$
\STATE Initialize best score $S_{\mathrm{best}} \leftarrow -\infty$

\FOR{$t = 1$ to $T$}

    \STATE Generate candidates target descriptions $\mathcal{C}^{(t)} = \{C_k^{(t)}\}_{k=1}^{K}$ \\ using $\mathcal{F}_{\mathrm{Executor}}(I_R, T_M, \boldsymbol{P}, F^{(t-1)})$

    \FOR{$k = 1$ to $K$}
        \STATE Compute $\mathrm{Rec}(T_M, C_k^{(t)})$, $s_{\mathrm{mod}}(k)$, and $s_{\mathrm{ref}}(k)$
        \STATE Compute the critic score $S_k$ using Eq.~\eqref{eq:critic_score}
    \ENDFOR

    \STATE $k^\ast \leftarrow \arg\max_k S_k$

    \IF{$S_{k^\ast} > S_{\mathrm{best}}$}
        \STATE $S_{\mathrm{best}} \leftarrow S_{k^\ast}$
        \STATE $Q \leftarrow C_{k^\ast}^{(t)}$
    \ENDIF

    \IF{$C_{k^\ast}^{(t)}$ satisfies the Critic's reliability criteria}
        \STATE Set final surrogate query $Q \leftarrow C_{k^\ast}^{(t)}$
        \STATE \textbf{break}
    \ELSE
        \STATE Generate structured feedback $F^{(t)}$ from the Critic evaluation
    \ENDIF

\ENDFOR

\STATE Rank all $I_i \in \mathbf{G}$ by cosine similarity $\cos(\boldsymbol{f}_\theta(I_i), \boldsymbol{f}_\phi(Q))$
\STATE \textbf{return} ranked gallery images

\end{algorithmic}
\end{algorithm} 

Algorithm~\ref{alg:pec_cir} summarizes the inference-time procedure of PEC-CIR. 
Given a reference image $I_R$ and a modification instruction $T_M$, the Planner constructs a structured plan, the Executor generates candidate target descriptions, and the Critic selects or refines candidates through iterative feedback.
The process terminates when a reliable candidate is obtained or the maximum number of iterations $T$ is reached.
The final query is then encoded by the text encoder $\mathbf{f}_{\phi}$ to retrieve gallery images.

%%%%%%%%%%%%%%%%%%%%% Experiments %%%%%%%%%%%%%%%%%%%%%
\section{Experiments}
\label{sec:experiments}
We evaluate PEC-CIR in a training-free zero-shot setting on the FashionIQ and CIRR benchmarks.
The experiments assess the effectiveness of the proposed Planner--Executor--Critic framework, including multi-candidate generation, Critic-based verification, and feedback-driven refinement.
We further conduct ablation studies to analyze the contribution of each component.

\subsection{Benchmarks and Evaluation Metrics}
We evaluate PEC-CIR on two representative benchmarks: FashionIQ, which focuses on fine-grained attribute modifications (e.g., color, style) in a controlled domain, and CIRR, which involves diverse everyday scenes requiring complex reasoning and scene-level transformations.
For FashionIQ, we report average Recall@10 and Recall@50 across three categories.
For CIRR, we report Recall@K (K=1, 5, 10, 50) on the full gallery and $\text{Recall}_\text{Subset}@\text{K}$ (K=1, 2, 3) on visually similar subsets to assess ranking precision under strong distractors.

\subsection{Implementation details}
\label{subsec:impl}
For the retrieval model, we utilize the frozen ViT-G/14 backbone \cite{cherti2023reproducible} from OpenCLIP, which was pre-trained on the LAION-2B dataset, with image and text representations extracted via $\boldsymbol{f}_\theta$ and $\boldsymbol{f}_\phi$, respectively.
Final ranking is performed using cosine similarity between the normalized query embedding and gallery features.
All Planner--Executor--Critic modules are implemented using Gemini 2.0 Flash \cite{gemini2flash}, which serves as the LLM backbone for structured planning, candidate generation, and Critic-based evaluation.
Existing training-free CIR methods use heterogeneous LLM/MLLM backbones and retrieval pipelines, making a fully normalized comparison across all baselines difficult.
For baselines that we directly reimplement, we use the same frozen ViT-G/14 retrieval backbone for final gallery ranking.
Results quoted from prior work are reported under their original experimental protocols and are marked as reported results.
We further analyze the sensitivity of PEC-CIR to the LLM backbone in Section~\ref{subsec:ablation_llm_backbones}.
The Planner--Executor--Critic loop is executed for a maximum of $T=2$ iterations, with the Executor generating $K=3$ candidates per round.
Hyperparameters for the Critic score are set to $w_{\mathrm{high}}=0.4$, $w_{\mathrm{med}}=0.25$, and $w_{\mathrm{low}}=0.1$.
To maintain numerical stability, the length regularization term $\Delta_{\mathrm{len}}$ is bounded within $[-0.05, 0.05]$, while the penalty terms $\Delta_{\mathrm{overlap}}$ and $\Delta_{\mathrm{prohibit}}$ are bounded within $[0, 0.2]$.
We adopt a strictly training-free protocol, without task-specific parameter updates, additional optimization, or synthetic training data generation.

\subsection{Main results}
\label{subsec:main_results}
\subsubsection{Evaluation on FashionIQ}
Table~\ref{tab:fiq_comparison} compares PEC-CIR with training-free and training-based zero-shot methods on FashionIQ.
PEC-CIR achieves the best performance among the compared training-free methods, with an average Recall@10 of 43.43\%, outperforming the strongest training-free baseline, MCoT-RE \cite{park2025mcot}, by 2.56\%.
Notably, PEC-CIR also outperforms the strongest training-based baseline in Table~\ref{tab:fiq_comparison}, FTI4CIR \cite{lin2024fine}, by 14.04\% in average Recall@10.
The consistent improvements across all categories indicate that the Planner--Executor--Critic loop improves query construction by enhancing instruction fidelity and reference preservation.

% ------------------------------------------------------------
% Table 1  FashionIQ comparison (ViT-G/14 backbone)
% ------------------------------------------------------------
\begin{table}[!t]
\captionsetup{width=\linewidth}
\caption{Comparison on FashionIQ validation using a ViT-G/14 backbone.
The best and second-best results in each column are shown in bold and underlined, respectively.}
\centering
\setlength{\tabcolsep}{1.85pt}
\renewcommand{\arraystretch}{0.98}
{
\begin{tabular}{lcccccccc}
\toprule
\multirow{2}{*}{Method} &
\multicolumn{2}{c}{Shirt} &
\multicolumn{2}{c}{Dress} &
\multicolumn{2}{c}{Toptee} &
\multicolumn{2}{c}{Avg.} \\
\cmidrule(lr){2-3}\cmidrule(lr){4-5}\cmidrule(lr){6-7}\cmidrule(lr){8-9}
& R@10 & R@50 & R@10 & R@50 & R@10 & R@50 & R@10 & R@50 \\
\midrule

\multicolumn{9}{l}{Methods requiring additional training or optimization} \\
\midrule
Pic2Word \cite{saito2023pic2word}
& 20.00 & 40.20 & 26.20 & 43.60 & 27.90 & 47.40 & 24.70 & 43.70 \\
SEARLE-XL \cite{agnolucci2025isearle}
& 20.48 & 43.13 & 26.89 & 45.58 & 29.32 & 49.97 & 25.56 & 46.23 \\
KEDs \cite{suo2024knowledge}
& 21.70 & 43.80 & 28.90 & 48.00 & 29.90 & 51.90 & 26.80 & 47.90 \\
LinCIR \cite{gu2023language}
& 20.92 & 42.44 & 29.10 & 46.81 & 28.81 & 50.18 & 26.28 & 46.49 \\
ContextI2W \cite{tang2024context}
& 23.10 & 45.30 & 29.70 & 48.60 & 30.60 & 52.90 & 27.80 & 48.90 \\
TSCIR \cite{wang2025mapping}
& \underline{24.14} & \underline{46.80} & \underline{31.01} & \underline{50.05} & \textbf{32.94} & \textbf{54.26} & 29.37 & 50.37 \\
FTI4CIR \cite{lin2024fine}
& \textbf{24.39} & \textbf{47.84} & \textbf{31.35} & \textbf{50.59} & \underline{32.43} & \underline{54.21} & \textbf{29.39} & \textbf{50.88} \\

\midrule
\multicolumn{9}{l}{Training-free zero-shot methods} \\
\midrule
CIReVL \cite{karthik2023vision}
& 33.71 & 51.42 & 27.07 & 49.63 & 35.80 & 56.14 & 32.19 & 52.36 \\
AutoCIR \cite{cheng2025generative}
& 36.36 & 55.84 & 26.18 & 47.69 & 37.28 & 60.38 & 33.27 & 54.63 \\
LCR \cite{sun2023training}
& 35.38 & 55.15 & 24.84 & 45.56 & 33.10 & 53.60 & 31.11 & 51.44 \\
MRA-CIR \cite{tu2025multimodal}
&  40.43 & 60.20 & 31.87 & 54.23 & 41.25 & 62.51 & 37.85 & 58.98 \\
OSrCIR \cite{tang2025reason} & 38.65 & 54.71 & 33.02 & 54.78 & 41.04 & 61.83 & 37.57 & 57.11 \\
WeiMoCIR \cite{wu2024training}
& 37.73 & 56.18 & 30.99 & 52.45 & 42.38 & 63.23 & 37.03 & 57.29 \\
MCoT-RE \cite{park2025mcot}
& \underline{42.35} & \underline{59.81} & \underline{34.51} & \underline{56.67} & \underline{45.74} & \underline{67.57} & \underline{40.87} & \underline{61.35} \\
\midrule
\textbf{PEC-CIR (Ours)}
& \textbf{45.14} & \textbf{63.49}
& \textbf{38.03} & \textbf{60.24}
& \textbf{47.12} & \textbf{70.09}
& \textbf{43.43} & \textbf{64.61} \\
\bottomrule
\end{tabular}
}
\label{tab:fiq_comparison}
\end{table}

\subsubsection{Evaluation on CIRR}
Table~\ref{tab:cirr_comparison} reports the performance on CIRR under the official evaluation protocol.
Among the compared training-free methods, PEC-CIR achieves the highest Recall@1 of 39.46\%, outperforming OSrCIR \cite{tang2025reason} by 2.20\%.
The improvement is also observed in Recall@10, where PEC-CIR achieves 81.02\%, compared with 77.33\% for OSrCIR.
In the visually similar subset evaluation, OSrCIR performs better on $R_{\mathrm{sub}}@1$ and $R_{\mathrm{sub}}@2$ by 1.12\% and 0.48\%, respectively, whereas PEC-CIR achieves the highest $R_{\mathrm{sub}}@3$ of 93.57\%.
These results indicate that PEC-CIR provides strong retrieval performance over the full gallery while remaining competitive on fine-grained subset ranking.
The improvements suggest that iterative verification and feedback-driven refinement help reduce constraint violations and hallucinated query content in complex compositional instructions.

% ------------------------------------------------------------
% Table 2  CIRR comparison (ViT-G/14 backbone)
% ------------------------------------------------------------
\begin{table}[t]
\caption{Comparison on CIRR test using a ViT-G/14 backbone.
The best and second-best results in each column are shown in bold and underlined, respectively.}
\centering
\setlength{\tabcolsep}{3.1pt}
\renewcommand{\arraystretch}{1.02}
{
\begin{tabular}{lccccccc}
\toprule
Method & R@1 & R@5 & R@10 & R@50 & $\text{R}_{sub}{@1}$ & $\text{R}_{sub}{@2}$ & $\text{R}_{sub}{@3}$ \\
\midrule

\multicolumn{8}{l}{Zero-shot CIR with additional training or optimization} \\
\midrule

PALAVRA \cite{cohen2022my}
& 16.62 & 43.49 & 58.51 & 83.95 & 41.61 & 65.30 & 80.94 \\
SEARLE \cite{baldrati2023zero}
& 24.00 & 53.42 & 66.82 & 89.78 & 54.89 & 76.60 & 88.19 \\
SEARLE-OTI \cite{baldrati2023zero}
& 24.27 & 53.25 & 66.10 & 88.84 & 54.10 & 75.81 & 87.33 \\
Slerp \cite{jang2024spherical}
& 28.19 & 55.88 & 68.77 & 88.51 & 61.13 & 80.63 & 90.68 \\
\midrule
\multicolumn{8}{l}{Training-free zero-shot methods} \\
\midrule
CIReVL \cite{karthik2023vision}
& 23.94 & 52.51 & 66.00 & 86.95 & 60.17 & 80.05 & 90.19 \\
LDRE \cite{yang2024ldre}
& 25.69 & 55.13 & 69.04 & 89.90 & 60.53 & 80.65 & 90.70 \\
AutoCIR \cite{cheng2025generative}
& 30.53 & 59.42 & 72.19 & \underline{91.47}
& 65.11 & 84.02 & 92.70 \\
OSrCIR \cite{tang2025reason}
& \underline{37.26} & \underline{67.25} & \underline{77.33} & --
& \textbf{69.22} & \textbf{85.28} & \underline{93.55} \\
LCR \cite{sun2023training}
& 32.82 & 61.13 & 71.76 & 85.28
& 66.63 & 78.58 & 82.68 \\
WeiMoCIR \cite{wu2024training}
& 31.04 & 60.41 & 72.27 & 90.89
& 58.84 & 78.92 & 89.64 \\
\midrule
\textbf{PEC-CIR (Ours)}
& \textbf{39.46} & \textbf{69.40} & \textbf{81.02} & \textbf{94.39}
& \underline{68.10} & \underline{84.80} & \textbf{93.57} \\
\bottomrule
\end{tabular}
}
\label{tab:cirr_comparison}
\end{table}

\subsection{Ablation Study}
\label{subsec:ablation}

\subsubsection{Effect of refinement depth}
\label{subsec:ablation_rounds}
We investigate the sensitivity of retrieval performance to the maximum refinement depth $T$, where $T=0$ denotes a direct single-pass generation baseline without the Planner--Executor--Critic loop.
As shown in Table~\ref{tab:ablate_rounds}, performance improves as $T$ increases and reaches its peak at $T=2$ across both benchmarks.
On CIRR, Recall@1 increases from 30.58\% at $T=0$ to 39.46\% at $T=2$, suggesting that iterative verification and feedback-driven refinement help correct initial generation errors.
Performance begins to saturate or decline beyond $T=3$.
This trend indicates that a limited refinement depth improves query construction, whereas excessive refinement introduces semantic drift and unnecessary verbosity, reducing the precision of the final query embedding.

\begin{table}[t]
\caption{Sensitivity to the maximum number of refinement rounds $T$.}
\centering
\setlength{\tabcolsep}{3.2pt}
\renewcommand{\arraystretch}{1.02}

\resizebox{\columnwidth}{!}{%
\begin{tabular}{c|ccccccc|cc}
\toprule
 & \multicolumn{7}{c|}{CIRR} & \multicolumn{2}{c}{FashionIQ} \\
$T$ & R@1 & R@5 & R@10 & R@50 & $\text{R}_{sub}{@1}$ & $\text{R}_{sub}{@2}$ & $\text{R}_{sub}{@3}$ & Avg. R@10 & Avg. R@50 \\
\midrule
0 & 30.58 & 60.75 & 73.23 & 91.93 & 63.90 & 82.70 & 91.49 & 41.62 & 62.05 \\
1 & 34.78 & 68.76 & 78.02 & \underline{93.98} & 66.93 & \underline{83.81} & \underline{92.03} & \underline{42.98} & \underline{63.87} \\
2 & \textbf{39.46} & \textbf{69.40} & \textbf{81.02} & \textbf{94.39} & \textbf{68.10} & \textbf{84.80} & \textbf{93.57} & \textbf{43.43} & \textbf{64.61} \\
3 & \underline{39.01} & \underline{68.82} & \underline{80.84} & 93.54 & \underline{67.32} & 83.32 & 91.57 & 43.10 & 64.12 \\
4 & 38.55 & 67.33 & 79.32 & 93.04 & 66.69 & 82.98 & 91.71 & 42.54 & 63.48 \\
5 & 33.52 & 63.17 & 75.62 & 91.95 & 64.63 & 82.93 & 92.01 & 41.10 & 61.92 \\
\bottomrule

\end{tabular}
}
\label{tab:ablate_rounds}
\end{table}

\subsubsection{Component ablation}
\label{subsec:ablation_components}
We evaluate the individual contributions of the Planner, Executor, and Critic modules to assess their effects on retrieval performance.
As shown in Table~\ref{tab:ablate_components}, removing each module or disabling iterative refinement consistently decreases recall.
Combining the Planner and Executor achieves 36.94\% Recall@1, showing a notable improvement over the single-pass generation baseline of 30.12\%.
The full PEC-CIR framework further improves Recall@1 to 39.46\%, highlighting the importance of Critic-based verification.
These results indicate that planning, candidate diversity, and verification make complementary contributions to the proposed architecture, improving retrieval performance through their combined effect.

\begin{table}[t]
\caption{Agent component ablation on CIRR.}
\centering
\setlength{\tabcolsep}{4.0pt}
\renewcommand{\arraystretch}{1.02}

\resizebox{\columnwidth}{!}{%
\begin{tabular}{lccc}
\toprule

Variant & R@1 & R@5 & R@10 \\
\midrule
Single-pass Generation (w/o Planner, w/o Critic) 
& 30.12 & 58.47 & 71.03 \\

Planner-guided Single-pass (w/o Critic) 
& 33.85 & 64.92 & 76.88 \\

Executor only (w/o Planner, w/o Critic) 
& 34.27 & 65.31 & 77.42 \\

Planner + Executor (w/o Critic) 
& 36.94 & 67.88 & 78.36 \\

PEC-CIR (w/o Refinement, $T=0$) 
& 30.58 & 60.75 & 73.23 \\

\textbf{PEC-CIR (Full)} 
& \textbf{39.46} & \textbf{69.40} & \textbf{81.02} \\

\bottomrule
\end{tabular}
}
\label{tab:ablate_components}
\end{table}

\subsubsection{Sensitivity to the number of candidates}
\label{subsec:ablation_k}
We analyze the impact of candidate diversity on retrieval performance by varying the number of generated candidates $K$ at each iteration.
As shown in Table~\ref{tab:k_sensitivity}, the framework achieves its highest performance at $K=3$ across both benchmarks.
Increasing $K$ from 1 to 3 expands the set of candidate descriptions and provides the Critic with a more diverse pool for selection.
However, performance declines beyond $K=4$, suggesting that an overly large candidate pool introduces noisy or less relevant descriptions into the selection process.
These results indicate that moderate candidate diversity is beneficial for verification, whereas excessive diversity reduces retrieval performance by introducing redundant or distracting linguistic information.

\begin{table}[t]

\caption{Sensitivity to the number of candidates $K$.}
\label{tab:k_sensitivity}
\centering
\setlength{\tabcolsep}{4.0pt}
\renewcommand{\arraystretch}{1.02}

\begin{tabular}{c|cccc|cc}
\toprule
 & \multicolumn{4}{c|}{CIRR} & \multicolumn{2}{c}{FashionIQ} \\
$K$ & R@1 & R@5 & R@10 & R@50 & Avg. R@10 & Avg. R@50 \\
\midrule
1 & 36.84 & 63.95 & 75.90 & 90.52 & 36.92 & 57.88 \\
2 & 38.72 & 66.31 & 78.08 & 91.23 & 38.18 & 59.43 \\
\textbf{3} & \textbf{39.46} & \textbf{69.40} & \textbf{81.02} & \textbf{94.39} & \textbf{43.43} & \textbf{64.61} \\
4 & 39.02 & 68.21 & 80.15 & 93.62 & 42.35 & 63.54 \\
5 & 37.58 & 65.47 & 77.84 & 92.11 & 40.96 & 61.98 \\
\bottomrule
\end{tabular}
\end{table}

\subsubsection{Effect of Critic scoring strategies}
\label{subsec:ablation_critic_variants}
We investigate the influence of different decision-making strategies within the Critic by comparing alternative candidate selection criteria.
This analysis includes a baseline without the Critic, a variant relying only on embedding similarity, a variant using unstructured LLM-based judgment, and the proposed multi-criteria scoring mechanism.
As shown in Table~\ref{tab:ablate_critic_variants}, selection strategies based on single or unstructured criteria yield lower performance than the full scoring strategy.
For example, the embedding similarity only variant achieves 34.12\% Recall@1, whereas the full PEC-CIR framework reaches 39.46\%.
These results indicate that explicit constraint modeling and score-based selection contribute to candidate selection in complex retrieval tasks.

\begin{table}[t]
\caption{Sensitivity to Critic selection strategies on CIRR.}
\centering
\setlength{\tabcolsep}{4.0pt}
\renewcommand{\arraystretch}{1.02}
\begin{tabular}{lccc}
\toprule
Critic selection strategy & R@1 & R@5 & R@10 \\
\midrule
No selection (w/o Critic)
& 30.58 & 60.75 & 73.23 \\

Critic with embedding similarity only 
& 34.12 & 64.38 & 77.01 \\

Critic with LLM-based judgment only 
& 35.46 & 65.02 & 77.94 \\

\textbf{Score-based Critic (Full)} 
& \textbf{39.46} & \textbf{69.40} & \textbf{81.02} \\
\bottomrule
\end{tabular}
\label{tab:ablate_critic_variants}
\end{table}

\subsubsection{LLM backbone analysis}
\label{subsec:ablation_llm_backbones}
To examine whether PEC-CIR depends on a specific LLM backbone, we additionally evaluate the same framework with GPT-4o-mini, Gemini 2.5 Flash Lite, and Qwen3-VL-8B-Instruct, as shown in Table \ref{tab:llm_comparison}.
Specifically, the average R@10, R@50 scores are 50.71, 70.25 with GPT-4o-mini, 50.18, 69.50 with Gemini 2.5 Flash Lite, and 50.33, 69.53 with Qwen3-VL-8B-Instruct.
Across the three LLMs, PEC-CIR achieves comparable average performance, with only small differences in both Recall@10 and Recall@50.
This consistency across LLM backbones supports the generality of the proposed Planner–Executor–Critic framework.

\begin{table}[!t]
\captionsetup{width=\linewidth}
\caption{Comparison on FashionIQ validation using a ViT-G/14 backbone.}
\centering
\setlength{\tabcolsep}{1.85pt}
\renewcommand{\arraystretch}{0.98}
{
\begin{tabular}{lcccccccc}
\toprule
\multirow{2}{*}{Method} &
\multicolumn{2}{c}{Shirt} &
\multicolumn{2}{c}{Dress} &
\multicolumn{2}{c}{Toptee} &
\multicolumn{2}{c}{Avg.} \\
\cmidrule(lr){2-3}\cmidrule(lr){4-5}\cmidrule(lr){6-7}\cmidrule(lr){8-9}
& R@10 & R@50 & R@10 & R@50 & R@10 & R@50 & R@10 & R@50 \\
\midrule
Gemini 2.0 Flash
& 45.14 & 63.49 & 38.03 & 60.24 & 47.12 & 70.09 & 43.43 & 64.61 \\
\midrule
GPT-4o-mini
& 52.16 & 68.99 & 43.28 & 66.09 & 56.86 & 75.83 & 50.71 & 70.25 \\
Gemini 2.5 Flash Lite
& 53.09 & 68.94 & 42.09 & 64.40 & 55.48 & 75.32 & 50.18 & 69.50 \\
Qwen3-VL-8B-Instruct
& 52.80 & 69.33 & 42.49 & 64.25 & 55.84 & 75.17 & 50.33 & 69.53 \\
\bottomrule
\end{tabular}
}
\label{tab:llm_comparison}
\end{table}

\subsubsection{Inference cost analysis}
To further analyze the practical cost of PEC-CIR, we report the inference cost on the FashionIQ validation set in Table \ref{tab:cost_analysis}.
We compare PEC-CIR with MCoT-RE, a representative single-pass training-free CIR method. 
MCoT-RE performs one MLLM request per query to generate two captions, whereas PEC-CIR requires 3.08, 3.09, and 3.18 LLM requests per query with GPT-4o-mini, Gemini 2.5 Flash Lite, and Qwen3-VL-8B-Instruct, respectively. 
This indicates that PEC-CIR introduces approximately three times more LLM requests than the single-pass MCoT-RE pipeline.

The measured token usage varies across LLM backbones because different providers use different tokenizers, chat templates, and usage accounting rules.
Nevertheless, the inference cost of PEC-CIR remains predictable and bounded by its fixed multi-stage design. 
Although PEC-CIR incurs additional inference cost compared with single-pass query generation, this cost is confined to the query-construction process and does not require task-specific training, fine-tuning, or retriever optimization. 

\begin{table}[!t]
\centering
\caption{Inference cost analysis on FashionIQ validation.}
\setlength{\tabcolsep}{4pt}
\renewcommand{\arraystretch}{0.98}
\begin{adjustbox}{width=\columnwidth}
\begin{tabular}{lcccc}
\toprule
Model (LLM) & \# Queries & Avg. req./query & Avg. input tok./query & Avg. output tok./query \\
\midrule
\multicolumn{5}{l}{MCoT-RE \cite{park2025mcot}} \\
\midrule
GPT-4o-mini
& 6,016 & 1.00 & 2,411.04 & 152.37 \\
Qwen3-VL-8B-Instruct
& 6,016 & 1.00 & 431.60 & 176.19 \\
Gemini 2.5 Flash Lite
& 6,016 & 1.00 & 188.33 & 71.08 \\
\midrule
\multicolumn{5}{l}{PEC-CIR (Ours)} \\
\midrule
GPT-4o-mini
& 6,016 & 3.08 & 9,722.12 & 262.83 \\
Qwen3-VL-8B-Instruct
& 6,016 & 3.18 & 1,463.50 & 335.64 \\
Gemini 2.5 Flash Lite
& 6,016 & 3.09 & 635.30 & 100.11 \\
\bottomrule
\end{tabular}
\end{adjustbox}
\label{tab:cost_analysis}
\end{table}

%%% appendix추가

\subsection{Qualitative Results}
\label{sec:qual}

\subsubsection{Qualitative success cases}
\label{app:qual_success}

\begin{figure}[!t]
\centering
\includegraphics[width=\textwidth]{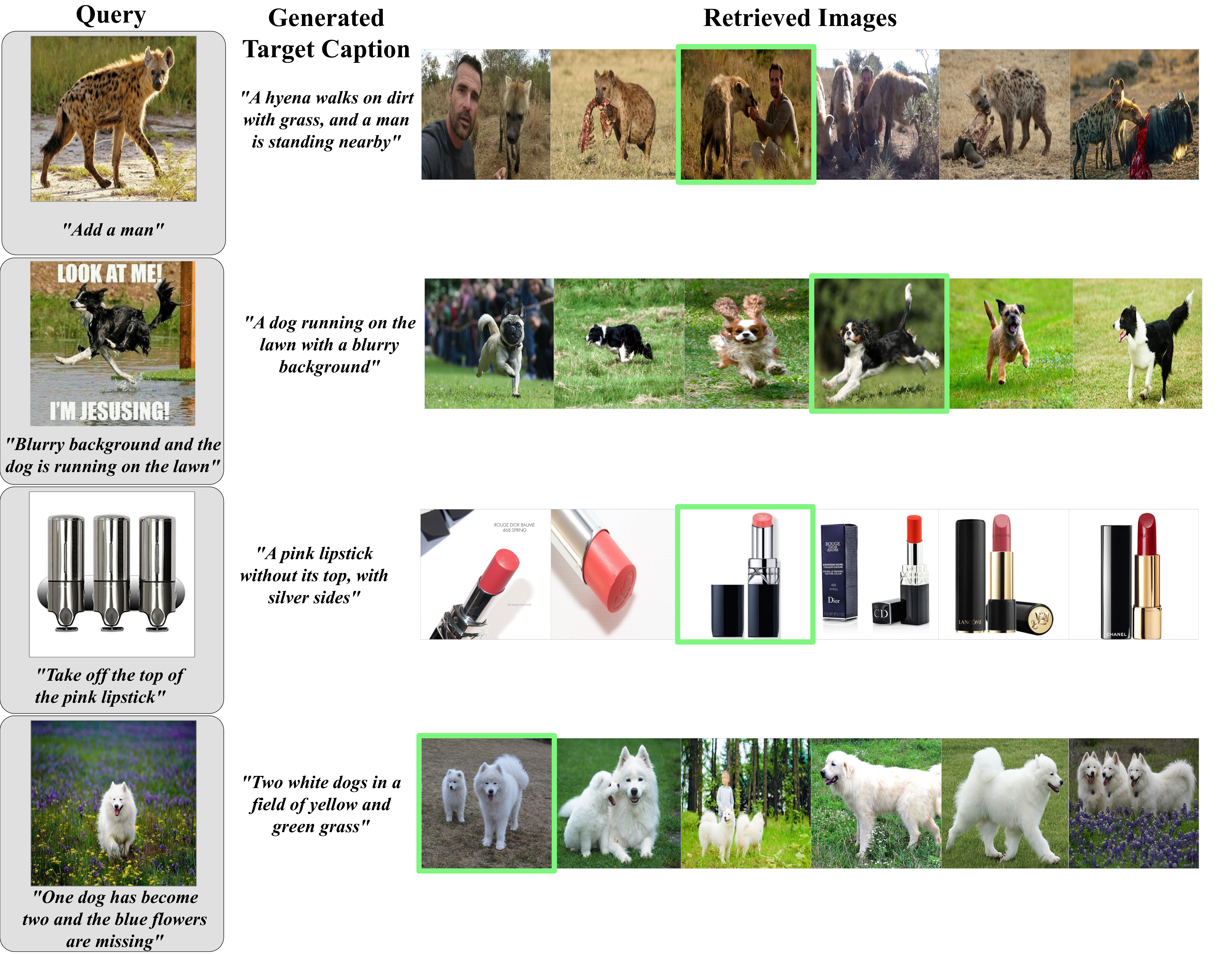}
\caption{Qualitative success cases of PEC-CIR. 
Each example displays the reference image, the modification instruction, the final generated query, and the retrieved target. 
The framework handles attribute and scene-level transformations while maintaining the core content of the reference image.}
\label{fig:qual_success}
\end{figure}

Figure~\ref{fig:qual_success} presents representative success cases of PEC-CIR under diverse modification instructions.
The examples cover several transformation types, including object addition, attribute modification, object count changes, and scene-level edits.
Across these cases, PEC-CIR retrieves target images that preserve the relevant visual content of the reference image while reflecting the requested textual modifications.
The generated queries capture key compositional cues in the instructions, such as added objects, changed attributes, and modified scene context.
These qualitative examples are consistent with the quantitative results, suggesting that the proposed inference pipeline supports query construction for multi-constraint instructions.

\subsubsection{Failure cases and error taxonomy}
\label{app:qual_failure}
\begin{figure}[!t]
\includegraphics[width=\textwidth]{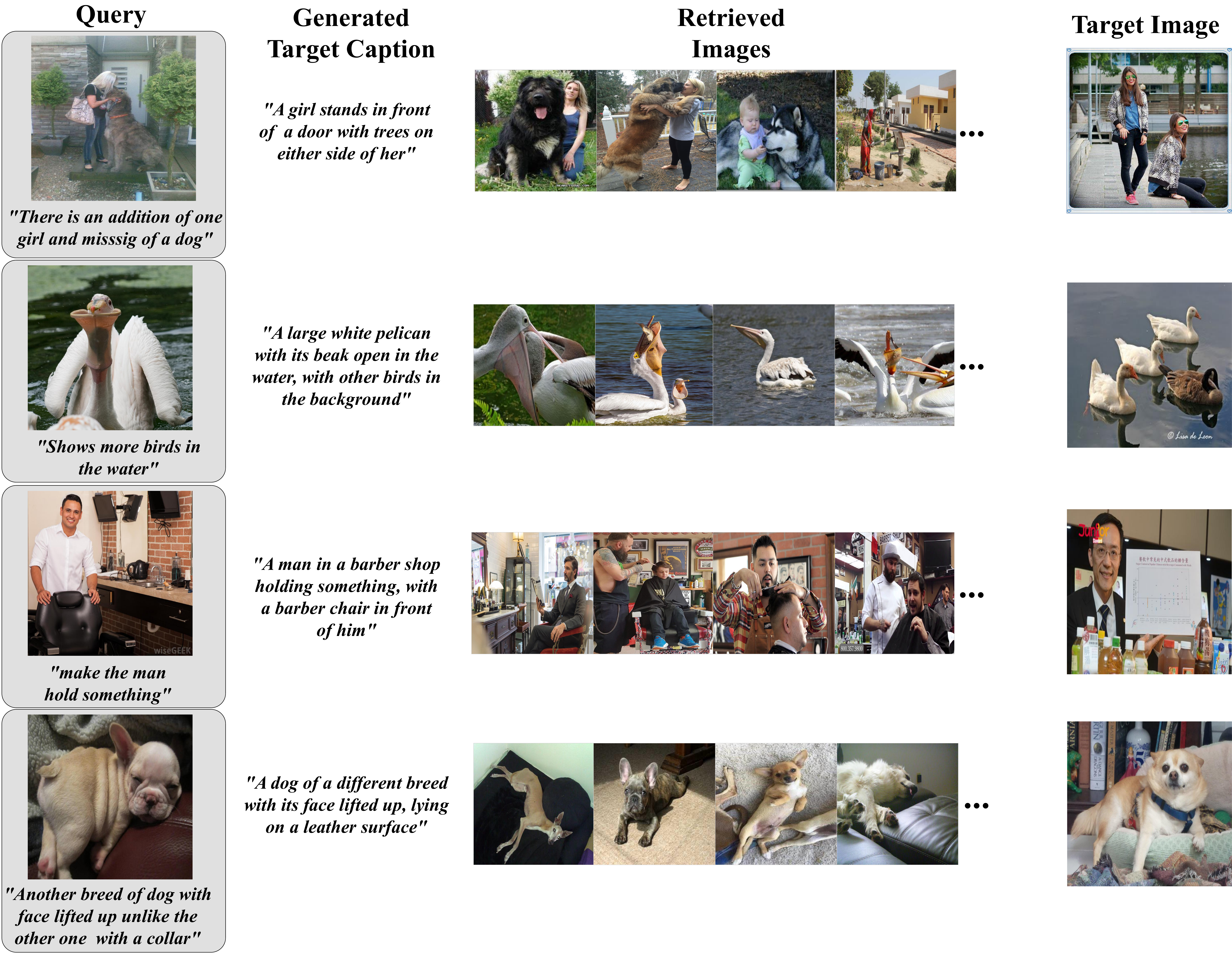}
\caption{Representative failure cases and error taxonomy of PEC-CIR. }

\label{fig:qual_failure}
\end{figure}

This section analyzes representative failure cases of PEC-CIR and groups them into three error types to identify the current limitations of the framework.
Figure~\ref{fig:qual_failure} presents examples from the CIRR validation set, where each instance includes the reference image, modification instruction, generated query, and the corresponding ground-truth target.

\textbf{Ambiguous or visually underspecified instructions.}
A frequent failure mode occurs when the modification text lacks specific visual cues.
Although such instructions are semantically coherent, they provide limited discriminative information for grounding the intended target within a frozen vision--language embedding space.
For instance, instructions such as ``make the man hold something'' or ``shows more birds in the water'' do not specify the exact object to be added or the concrete visual transformation required.
PEC-CIR therefore generates plausible target descriptions, but the ambiguity of the instruction leads to uncertainty during candidate selection and retrieval.
These failures mainly arise from limited instruction specificity, which makes it difficult to distinguish the ground-truth target from visually similar alternatives.

\textbf{Retrieval misalignment despite semantic plausibility.}
The second failure type involves cases where the selected candidate is semantically plausible but does not rank the ground-truth target highly.
In these instances, the generated query reflects the modification text and preserves relevant reference content, but it does not emphasize the visual cues most discriminative for retrieval.
This indicates a gap between linguistic plausibility and retrieval effectiveness in a training-free setting.
Even a semantically reasonable description may be insufficient to separate the target image from visually similar candidates in the frozen embedding space.

\textbf{Imbalance between preservation and modification.}
The third failure type is characterized by excessive preservation of the reference image, leading to insufficient emphasis on the requested modification.
This issue typically arises in scene-level edits or instructions involving multiple interacting constraints.
Although the Planner estimates preservation strength and the Critic evaluates reference consistency, the generated query sometimes remains biased toward the reference image identity.
As a result, the query partially reflects the instruction but lacks sufficient emphasis on the modified attributes required for accurate retrieval.
This limitation highlights the difficulty of balancing preservation and modification without direct feedback from the retrieval model.

\subsubsection{Analysis of Intermediate Outputs}
\label{app:intermediate_outputs}
\begin{figure}[!t]
\centering
\includegraphics[width=\textwidth]{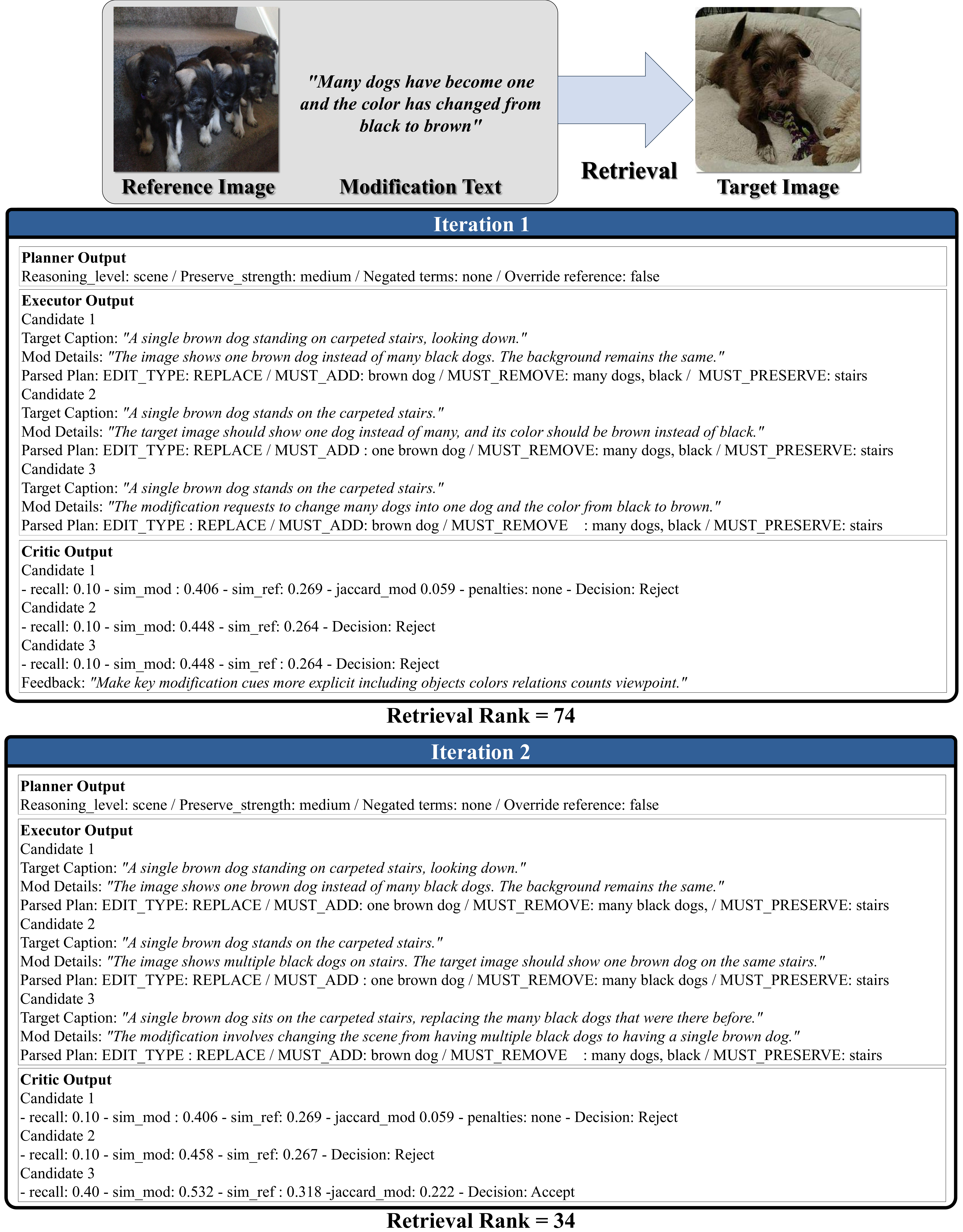}
\caption{Intermediate outputs of PEC-CIR across iterations. 
The figure visualizes the constraints derived by the Planner, the candidates generated by the Executor, and the corresponding evaluations from the Critic.}
\label{fig:intermediate_outputs}
\end{figure}

We analyze the intermediate outputs of PEC-CIR to illustrate the structured inference process involving planning, generation, verification, and feedback.
PEC-CIR operates through distinct stages in which candidate target descriptions are progressively refined across iterations.
Figure~\ref{fig:intermediate_outputs} visualizes this process using an example from the CIRR dataset.

\textbf{Constraint extraction by the Planner.}
Given a reference image and the instruction ``Many dogs have become one and the color has changed from black to brown'', the Planner converts the instruction into structured constraints.
The module classifies the instruction as a scene-level modification and identifies a replacement operation involving both object count and attribute changes.
Accordingly, the preservation strength is set to a medium level, and reference override is disabled because the instruction requires maintaining the core visual context.
The resulting plan provides a shared constraint structure that guides both candidate generation and Critic-based verification.

\textbf{Iterative candidate generation by the Executor.}
Conditioned on the Planner constraints, the Executor generates three candidate target descriptions at each iteration.
In the first iteration, the candidates reflect the color change but do not sufficiently emphasize the transition from multiple dogs to a single dog.
This omission weakens the representation of the core modification cues.
After receiving feedback from the Critic, the Executor generates revised candidates in the second iteration with clearer articulation of the ``many-to-one'' count change and the ``black-to-brown'' attribute replacement.
The revised candidates focus on the identified constraints without introducing unrelated interpretations.

\textbf{Verification and feedback by the Critic.}
The Critic evaluates each candidate using the criteria described in Section~\ref{subsec:critic}, including instruction coverage, reference consistency, prohibited-term compliance, and the composite score.
In the first iteration, the selected candidate is regarded as unreliable due to insufficient emphasis on key modification cues.
The Critic therefore provides feedback requesting explicit mention of the object count and color transitions.
In the second iteration, the revised candidates better reflect the required modification cues, and the selected candidate achieves a higher retrieval rank than the initial query.
The refinement process terminates after the Critic accepts this candidate as the final surrogate query.

This example illustrates how PEC-CIR uses intermediate outputs to structure the inference process.
The Planner defines explicit constraints, the Executor generates candidate realizations of these constraints, and the Critic evaluates the candidates while providing corrective feedback.
The improvement in retrieval rank across iterations suggests that Critic-guided refinement contributes to constructing a more effective final query representation.

%%%%%%%%%%%%%%%%%%%%% Conclusion %%%%%%%%%%%%%%%%%%%%%
\section{Conclusion}
This paper identifies three structural challenges in training-free zero-shot composed image retrieval: decoupled cross-modal processing, stochastic generation volatility, and limited discriminative oversight.
To address these challenges, we introduced PEC-CIR, a structured reasoning framework that replaces single-pass query generation with an iterative planning, execution, and verification pipeline.
By introducing a Planner--Executor--Critic pipeline, PEC-CIR improves the stability of query construction through planning, multi-candidate generation, and iterative verification.
Experimental results on FashionIQ and CIRR demonstrate consistent improvements over existing training-free approaches, while qualitative analyses show its effectiveness in handling complex modification instructions.

Despite these improvements, several limitations remain.
Ambiguous instructions and visually subtle modifications continue to pose challenges for stable interpretation.
In addition, the Planner extracts prohibited terms from explicitly stated exclusion expressions, so indirect instructions with implied negative meanings are not always converted into prohibited terms. 
This limitation affects cases where the intended modification implicitly excludes certain visual conditions without directly naming them.
To address this issue, future extensions of the Planner will incorporate semantic expansion based on commonsense and entailment cues, so that implicit exclusions can be represented as soft constraints during candidate verification. 
Another limitation is that the Critic relies on predefined evaluation criteria, which may not fully capture subjective user preferences or task-specific retrieval intents.
Moreover, although our LLM backbone analysis shows that PEC-CIR achieves comparable performance across GPT-4o-mini, Gemini 2.5 Flash Lite, and the open-source Qwen3-VL-8B-Instruct, the framework still inherits common limitations of LLM-based training-free methods.
In particular, its outputs can be affected by decoding sensitivity, prompt sensitivity, and model version updates, especially when API-based LLM backbones are used.

Future work will explore the integration of explicit user feedback to support more adaptive refinement.
Dynamically updating the verification criteria based on user interaction is a promising direction for improving the flexibility and robustness of agentic retrieval systems.

\section{Acknowledgments}
This work was partly supported by Institute of Information \& Communications Technology Planning \& Evaluation (IITP) under the artificial intelligence graduate school program (Korea University) (No. RS-2019-II190079) and artificial intelligence star fellowship support program to nurture the best talents (IITP-2026-RS-2025-02304828) grant funded by the Korea government (MSIT).

% \end{thebibliography}
\end{document}